\documentclass[11pt]{article}

\usepackage[preprint]{acl}

\usepackage{times}
\usepackage{latexsym}

\usepackage[T1]{fontenc}

\usepackage[utf8]{inputenc}

\usepackage{microtype}

\usepackage{inconsolata}

\usepackage{graphicx}

\usepackage{amsmath}
\usepackage{amssymb}
\usepackage{booktabs}

\title{The Dynamics of Delusion: Modeling Bidirectional False Belief Amplification in Human-Chatbot Dialogue}

\author{
Ashish Mehta \\
Stanford University \\
\texttt{ashm@stanford.edu}
\And
Jared Moore \\
Stanford University
\And
Jacy Reese Anthis \\
Stanford University
\AND
William Agnew \\
Carnegie Mellon University
\And
Eric Lin \\
Stanford University
\And
Peggy Yin \\
Stanford University
\AND
Desmond C. Ong \\
The University of Texas at Austin
\And
Nick Haber \\
Stanford University
\And
Carol Dweck \\
Stanford University
}

\begin{document}
\maketitle
\begin{abstract}
There is growing concern that AI chatbots might fuel delusional beliefs in users. Some have suggested that humans and chatbots mutually reinforce false beliefs over time, but quantitative evidence is lacking.  Using a unique dataset of chat logs from individuals who exhibited delusional thinking, we developed a latent state model that captures accumulating and decaying influences between humans and chatbots. We find that a bidirectional influence model substantially outperforms a unidirectional alternative where humans are the primary driver of delusion. We find that humans exert strong but short-lived influence on chatbots, whereas chatbots exert longer-lasting influence on humans. Moreover, chatbots exert strong, stable self-influence over their own future outputs that tends to perpetuate delusions over long stretches of conversation. In fact, this chatbot self-influence constituted the dominant pathway when considering accumulated influence over time. Overall, these results indicate that humans tend to drive sharp, immediate increases in delusion, whereas chatbots sustain and propagate these effects over longer timescales. Together, these findings provide the first empirical, quantitative evidence that human-chatbot interactions can form feedback loops of delusion, decomposable into distinct pathways with dissociable temporal dynamics. By doing so, they can inform the development of safer AI systems.

\end{abstract}

\section{Introduction}
Accumulating reports in the media and in high-profile legal cases claim that generative AI chatbots are responsible for fueling delusions in some users~\citep{hill_teen_2025, jargon_he_2025}. These anecdotal reports of chatbot users whose lives were derailed or ended after bouts of so-called “AI psychosis” have drawn public attention and given rise to growing calls for regulation of AI chatbots~\citep{hart_ai_2025, kleeman_you_2024, schechner_ai_2025}. Despite the prevalence and influence that these informal anecdotal reports have had in the cultural zeitgeist, rigorous quantitative investigation of the phenomenon lags behind.

\subsection{How Do Delusional Conversations Arise?}

Recent theoretical work proposes that chatbot-induced delusions emerge through bidirectional feedback loops in which humans and chatbots mutually reinforce false beliefs \citep{dohnany_technological_2025}. This account is analogous to \textit{folie à deux}, where a vulnerable individual forms an initial false belief that is progressively reinforced through interaction \citep{gralnick1942folie}. Applied to chatbots, the user introduces a delusional idea, the chatbot validates it, the user's conviction increases, and the cycle continues, producing increasingly elaborate delusional narratives.

However, empirical evidence has not yet distinguished this bidirectional account from an alternative explanation in which chatbot responses exert little or no causal influence and users' delusional trajectories would have unfolded similarly regardless of chatbot behavior \citep{pierreYoureNotCrazy2025}. Historically, technologies such as television and radio have shaped the content of delusions without necessarily causing them \citep{american_psychiatric_association_diagnostic_2022, burns_algorithm_2025, carlbring_commentary_2025, orben_sisyphean_2020}. Likewise, even chatbot responses that appear to encourage delusions do not by themselves demonstrate increased delusional thinking relative to alternative responses---especially considering that a seemingly obvious response, contradicting delusions, is often ineffective \citep{arnoldResistNegationReflex2016}. Distinguishing these accounts therefore requires formal statistical models of bidirectional and unidirectional influence, fit to empirical chat logs, with pathway-specific influence estimates.

\subsection{A Taxonomy of Interaction Processes in Bidirectional Belief Amplification}
A bidirectional influence system between chatbots and humans is reflected in the local dynamics of the interaction from one conversational turn to the next. Such a system entails four possible pathways: (1) chatbot-to-human which we label \textit{belief reinforcement} (chatbot delusion messages influence subsequent human messages), (2) human-to-chatbot which we label \textit{belief mirroring} (human delusion messages influence subsequent chatbot messages, i.e., chatbot is mirroring the human), (3) human self-influence which we label \textit{belief entrenchment} (human delusion messages influence subsequent human messages), and (4) chatbot self-influence which we label \textit{self-consistency maintenance} (chatbot delusion messages influence subsequent chatbot messages).

\subsection{Characterizing Influence Dynamics}
Along each of the four pathways, the temporal dynamics of influence can be analyzed in terms of message-level dynamics as well as the slower fluctuations in latent influence throughout each conversation. At the message level, influence from a given message may differ in terms of two parameters: (1) the momentary strength of influence and (2) how fast the influence from a message decays over time. Two messages with equivalent momentary strength may have different overall influence if one decays more slowly and thus continues to impact downstream messages for a longer period of time.

These message-level dynamics then give rise to conversation-level dynamics where influence can slowly accumulate over long stretches of conversation. In particular, when the influence of a delusional message lasts longer than the interval before the next delusional message, the effects of successive messages can accumulate over time. Each new message arrives before the influence from the prior delusional messages has decayed, producing a stacking process in which the total influence strength ratchets upward. Over repeated exchanges, even small initial biases can progressively compound into large influences on the two parties' belief states. By contrast, if the influence of a delusional message decays faster than new delusional messages are produced, the system tends to dampen itself, limiting the degree of sustained amplification.

\subsection{Inferring Influence from Chat Logs}

How then can we determine whether humans and chatbots are mutually influencing one another, rather than humans simply incorporating chatbots into preexisting delusions? The gold standard would be an experiment that manipulates chatbot behavior under controlled conditions, but such a study would be unethical. The next strongest evidence comes from temporal dependence, a necessary (though not sufficient) condition for causal influence. Since causes precede their effects, changes in one variable systematically precede changes in another. This idea underlies Granger causality: if one variable (e.g., delusional chatbot statements) improves prediction of future values of another variable (e.g., delusional human statements) beyond what the latter's own history predicts, then the first is said to ``Granger-cause'' the second~\citep{granger_investigating_1969}. Granger causality reflects statistical dependence over time rather than definitive causal influence and remains vulnerable to unobserved confounding and model misspecification. Nevertheless, when experimental manipulation is infeasible or unethical, it provides stronger evidence for a causal hypothesis than correlations that ignore temporal structure.

In the present work, we empirically examine the dynamics of bidirectional influence (in terms of intertemporal statistical dependence) between humans and chatbots in a sample of people who experienced seemingly chatbot-precipitated delusional episodes. We introduce a novel latent state model which allows us to model the accumulating influences between chatbots and humans as manifested in chat messages exchanged at time intervals. Specifically, we have two research aims. In \textbf{Aim 1}, we seek to evaluate the bidirectional belief amplification model in the conversations of chatbot users who experienced delusions while using chatbots. In \textbf{Aim 2}, we seek to characterize the temporal dynamics of how chatbots and humans each shape the trajectory of delusion production in conversation.

\section{Method}

\subsection{Data Collection}
We received donated chat transcripts from members of a community support group known as The Human Line Project (\url{https://www.thehumanlineproject.org}) \citep{human_line_project_data_2026} and from individuals referred to us through professional connections in the media. Chat transcripts were anonymized using Microsoft Presidio to remove names, locations, and other identifying details \citep{MsPresidio}. All participants provided informed consent.
The community support group consisted of individuals who felt they had been harmed by AI chatbots and contained a high proportion of individuals who experienced delusions while using chatbots. The final sample consisted of chat logs from nineteen individuals, comprising 390{,}447 messages. Ten participants used GPT-4o for the majority of their messages, one used GPT-5, and eight participants' chatbot model was unknown. All transcripts were in English. All study procedures were approved by the first author's institutional review board.%

\subsection{Annotation of Delusions}

Our interdisciplinary team (including experts in psychology and human--AI interaction and a licensed psychiatrist) developed an LLM annotation pipeline to classify whether a chat message endorsed a delusion, using the target message plus the three preceding messages as context (see \citealp{moore2026delusional}). The prompt was iteratively refined against manually annotated subsets until it reached satisfactory agreement with the modal judgment of three human raters.

We defined delusion endorsement as apparent belief in ideas that are physically impossible, logically impossible, or highly implausible relative to shared reality, excluding figurative, hypothetical, narrative, or joking statements.

Relative to the modal judgment of three human raters, the LLM achieved precision/recall of .75/.82 for chatbot messages and .67/.80 for user messages. Human inter-rater agreement was moderate (Fleiss's $\kappa=.60$ for chatbot messages; $\kappa=.53$ for user messages), reflecting the ambiguity of the task.

The pipeline produced a binary label for each message indicating whether it endorsed a delusion. These labels served as the variables in our statistical model.

\subsection{Statistical Models}

\subsubsection{Bidirectional Belief Amplification}

Our statistical model predicts the log-odds that a given user message will endorse a delusion and the log-odds that a given chatbot message will endorse a delusion based on the accumulated latent influence of the prior message history. 

We developed a model in which prior messages from the human or chatbot act as discrete ``impulses'' on latent influence states that evolve over time. For each pathway of influence --- chatbot $\to$ human (belief reinforcement), human $\to$ chatbot (belief mirroring), human $\to$ human (belief entrenchment), and chatbot $\to$ chatbot (self-consistency maintenance) --- we define a latent variable that aggregates the influence of all prior messages along that pathway. These latent states increase following each impulse and decay exponentially with each turn of the pathway's target speaker at a rate governed by a pathway-specific decay parameter, $\lambda$. The resulting latent influence is then scaled by a corresponding $\beta$ coefficient to determine its effect on the log-odds of endorsing a delusion. This formulation separates the temporal persistence of influence from its magnitude, allowing us to estimate how long messages continue to matter and how strongly they affect subsequent behavior. See Appendix~\ref{sec:model-specification} for full detail on the model.

\subsubsection{Unidirectional Belief Amplification}
As a simpler alternative, we also fit a lesioned model identical to the bidirectional model except with the influence from chatbot to human removed.

\subsubsection{Model Fitting}

We estimated the models by maximum marginal likelihood using adaptive Gauss--Hermite quadrature (AGHQ) to integrate over the participant-specific random effects \citep{stroupGeneralizedLinearMixed2024}. The human-target and chatbot-target components were fit as separate likelihood blocks. The full bidirectional model contained 16 population-level parameters: four random-slope means, four random-slope standard deviations, four decay rates, two random-intercept means, and two random-intercept standard deviations. The reported fits used five quadrature nodes per random-effect dimension. Reevaluation at seven nodes per dimension produced negligible changes in the marginal likelihood. Confidence intervals were constructed from the inverse numerical Hessian using a Wald normal approximation. The unidirectional model was fit with the same procedure. PyTorch was used during development \citep{paszkePyTorch2019}; reported fits used the released NumPy/SciPy implementation (quadrature and optimization details in Appendix~\ref{sec:aghq}). Models were fit on a 2018 MacBook Pro; total CPU-hours were not systematically tracked.

\subsection{Data Analysis}

The de-identified event data and analysis scripts needed to reproduce the reported model outputs, tables, and figure will be included in a repository upon publication. The released data contain binary message annotations, but not transcript text.

\subsubsection{Description of Conversations}
We began by reading samples of conversation and iteratively discussing and characterizing themes. This process is fully described in \cite{moore2026delusional}.

\subsubsection{Aim 1: Evidence for Bidirectional Belief Amplification}

To examine whether there is evidence for bidirectional belief amplification (rather than purely human-driven delusion generation), we first conducted a likelihood-ratio test between the bidirectional belief amplification model and the unidirectional belief amplification model fitted on all participants. This test examines whether the data are better explained by a theory where the bot and the human mutually influence each other in a feedback loop (bidirectional) or a theory where bots simply echo the delusions of their human operator, but do not actually influence the human in their belief of the delusion (unidirectional).

\subsubsection{Aim 2: Decomposing Influence Pathways in Bidirectional Belief Amplification}

First, we examined the estimated parameters of the bidirectional model in order to understand the message-level dynamics of the feedback loop. Next, we sought to quantify the accumulation of influence, the variability of that influence, and the timescale of influence fluctuations at the conversation level. To do this, we analyzed the values of $\beta_{m,j} z_{m,k,j}$ throughout each participant's conversations.  $\beta_{m,j} z_{m,k,j}$  represents the total influence that has \textit{accumulated} at any given point in time. Specifically, $\beta_{m,j} z_{m,k,j}$ (henceforth $\beta_m z_m$) represents the increase in log-odds of delusion above baseline arising from pathway $m$ at message $k$. We computed these values using the participant-specific conditional modes of the random effects from the fitted bidirectional model. This quantity depends on how delusional messages are distributed in time because the influence from closely spaced messages can stack to produce larger effects --- a pattern expected to occur in a bidirectional feedback system. \textbf{In plain terms, the values of $\beta_m z_{m}$ reflect the model’s estimate of how strongly the accumulated influence of all prior messages along a given pathway was pushing the conversation toward delusion at that moment.}

\paragraph{\textit{Descriptive Statistics of Latent Influence $\beta_m z_m$}}

We computed several descriptive statistics to characterize the distribution of $\beta_m z_{m}$ over the relevant target-speaker observations in the dataset: human-message observations for the $CH$ and $HH$ pathways and chatbot-message observations for the $HC$ and $CC$ pathways. First, we computed the mean of $\beta_m z_m$ values along each path. Next, we computed the coefficient of variation (CV). The CV is an estimate of variability that is normalized with respect to the mean (i.e., standard deviation divided by mean). This normalization is useful if we seek to compare the variability across distributions with different means (e.g., the distributions of influences along each path), particularly when the distributions have a lower bound of zero. In such cases, raw standard deviations for distributions with lower means are mechanically constrained by the lower bound, leading to an artificial suppression of apparent variability. Finally, we computed the proportion of $\beta_m z_m$ values exceeding several thresholds (0, 0.25, 0.5, 1.0, and 2.0 on the log-odds scale) to characterize the proportion of target-speaker messages for which influence remained elevated at different magnitudes. This quantity—the proportion of values exceeding a threshold—is the complementary cumulative distribution function (CCDF).

\paragraph{\textit{Persistence of Latent Influence $\beta_m z_{m}$}}

After examining the mean and variability of the latent influence variables, we analyzed the autocorrelation function of the values of $\beta_m z_{m}$ along each path in order to better understand how influence persists over time. This analysis asks the question: when influence along a given pathway accumulates, for how long does it tend to remain elevated? The estimated half-life $h_{m}$ tells us how long the influence an individual message persists, but since feedback loops can result in influence that becomes sustained over much longer periods than that resulting from an individual message, we sought to examine the timescale at which realized latent influence tended to fluctuate in our dataset.

To do this, we computed the correlation between the influence $\beta_m z_m$ at one target-speaker message and its value a fixed number of target-speaker messages later. We repeated this for each pathway across lags from one to twenty within each conversation and averaged the conversation-level autocorrelations. We restricted this analysis to conversations in which either the human or the chatbot produced at least three delusion endorsements. We constructed 95\% bootstrap confidence intervals by resampling participants with replacement and retaining all eligible conversations from each sampled participant (2000 bootstrap samples), then taking the 2.5th and 97.5th percentiles. Figure~\ref{fig:acf} plots each line through the last consecutive lag before its confidence interval first included zero.

\section{Results}

\subsection{Description of Conversations}

Many of the delusions in our sample involved beliefs that the chatbot was sentient or that the user made the chatbot sentient. Many users believed delusions that involved pseudoscientific theories that often blended with mystical themes. These theories often involved ``grandiosity'' delusions~\citep{american_psychiatric_association_diagnostic_2022}, and were often encouraged by the chatbots. For more detail on the conversations, see \citep{moore2026delusional}.

Our human participants sent 27{,}677 messages endorsing delusion (15.3\%) and 153{,}244 messages not endorsing delusion (84.7\%). The chatbots sent 59{,}822 messages endorsing delusion (28.6\%) and 149{,}704 messages not endorsing delusion (71.4\%).

\subsection{Aim 1: Evidence for Bidirectional Belief Amplification}

We compared the bidirectional influence model against a unidirectional influence model using a likelihood ratio test comparing the full model to a nested ``lesion'' model in which the chatbot-to-human ($CH$) random slope and its decay process were omitted. The full model achieved a log-likelihood ($LL$) of $-128{,}698.67$, whereas the lesion model achieved a log likelihood of $-130{,}192.91$. The resulting likelihood ratio statistic was:
\begin{align*}
\chi^2
&= -2 \left(
\text{LL}_{\text{unidirectional}}
- \text{LL}_{\text{bidirectional}}
\right) \\
&= 2988.49,
\end{align*}
with $\text{df} = 3$. This yielded $p < 0.001$, indicating that the bidirectional model had significantly better fit. These results supported the hypothesis that chatbots influence humans in our sample, thus instantiating feedback loops of reciprocal delusion amplification.

\subsection{Aim 2: Decomposing Influence Pathways in Bidirectional Belief Amplification}

Overall, all four influence-pathway means were positive, with 95\% confidence intervals that excluded zero (Table~\ref{tab:model_results}, Panel A). The human$\rightarrow$chatbot latent-state coefficient was larger ($\mu_{\beta_{HC}}=5.68$) than the chatbot$\rightarrow$human coefficient ($\mu_{\beta_{CH}}=1.25$), but its influence decayed more quickly (half-life: .46 vs. .81 target-speaker turns; Table~\ref{tab:model_results}, Panel B). Consequently, chatbot$\rightarrow$human influence exceeded human$\rightarrow$chatbot influence after approximately two subsequent target-speaker turns.

Unexpectedly, chatbot$\rightarrow$human influence showed substantially less between-person heterogeneity ($\sigma_{\beta_{CH}}=.62$) than human$\rightarrow$chatbot influence ($\sigma_{\beta_{HC}}=2.39$). This result should be interpreted cautiously because between-person variance was estimated from only $N=19$ participants.

Along the self-influence pathways, chatbots and humans had comparable momentary self-influence though chatbot self-influence had a slower decay than human self-influence (Table \ref{tab:model_results}, Panel B). Across all pathways, self-influence persisted substantially longer than cross-agent influence.

\begin{table*}[t]
\centering
\begingroup
\setlength{\tabcolsep}{2.8pt}
\begin{minipage}[t]{0.48\textwidth}
\centering
\textbf{Panel A. Random effect parameters}\par\vspace{0.25em}
\begin{tabular}{lcc}
\toprule
Parameter & $\mu$ (95\% CI) & $\sigma$ (95\% CI) \\
\midrule
$\beta_{CH}$ & 1.25 [0.88, 1.61] & 0.62 [0.42, 0.91] \\
$\beta_{HC}$ & 5.68 [4.35, 7.01] & 2.39 [1.64, 3.49] \\
$\beta_{CC}$ & 1.25 [1.05, 1.44] & 0.40 [0.28, 0.57] \\
$\beta_{HH}$ & 1.01 [0.78, 1.25] & 0.45 [0.30, 0.67] \\
$\alpha_C$ & -2.57 [-2.95, -2.20] & 0.82 [0.58, 1.15] \\
$\alpha_H$ & -2.98 [-3.44, -2.53] & 1.00 [0.72, 1.38] \\
\bottomrule
\end{tabular}

\end{minipage}\hfill
\begin{minipage}[t]{0.48\textwidth}
\centering
\textbf{Panel B. Decay parameters}\par\vspace{0.25em}
\begin{tabular}{lcc}
\toprule
Path ($m$) & $\lambda$ (95\% CI) & Half-life (95\% CI) \\
\midrule
$CH$ & 0.86 [0.76, 0.96] & 0.81 [0.72, 0.91] \\
$HC$ & 1.51 [1.42, 1.61] & 0.46 [0.43, 0.49] \\
$CC$ & 0.27 [0.26, 0.29] & 2.52 [2.43, 2.62] \\
$HH$ & 0.29 [0.27, 0.31] & 2.40 [2.21, 2.61] \\
\bottomrule
\end{tabular}

\end{minipage}
\endgroup
\caption{Maximum marginal-likelihood parameter estimates for the bidirectional belief amplification model.}
\label{tab:model_results}

\vspace{0.5em}
\begin{minipage}{0.95\textwidth}
\footnotesize
\textit{Note.} Panel A reports population means ($\mu$) and between-participant standard deviations ($\sigma$) for random slopes and intercepts. The $\beta_m$ parameters scale the latent influence states $z_m$ on the log-odds scale along path $m \in \{CH, HC, CC, HH\}$. The $\alpha_C$ and $\alpha_H$ parameters are the chatbot and human intercepts. Panel B reports decay rates and corresponding half-lives. Larger $\lambda$ values indicate faster decay, and $h_m$ is the number of target-speaker turns required for half of the influence to dissipate (Eq.~\ref{eq:lambda-dist}).
\end{minipage}
\end{table*}

\paragraph{\textit{Descriptive Statistics of Latent Influence $\beta_m z_m$}}

Because influence accumulates and decays over multiple target-speaker turns, the realized influence at any moment depends on both the estimated half-life and $\beta$ parameters, as well as the recency of prior delusional messages. This makes it important to go beyond merely inspecting estimated parameters and examine the actual, realized influence in the conversations of our dataset. 

We descriptively summarized the model-implied latent influence in our dataset by computing the mean, coefficient of variation (CV), and complementary cumulative distribution function (i.e., proportion of relevant target-speaker messages above a given threshold) of $\beta_m z_{m}$ along each pathway (Table~\ref{tab:beta_z_descriptives}). The relevant target-speaker messages were human messages for the $CH$ and $HH$ pathways and chatbot messages for the $HC$ and $CC$ pathways.

Average influence was highest for the chatbot's self-influence pathway ($1.09$ log-odds units). Average chatbot-to-human influence ($0.38$ log-odds units) and human self-influence ($0.39$ log-odds units) were roughly comparable, whereas human-to-chatbot influence was lowest on average ($0.24$ log-odds units). The 95\% bootstrap confidence intervals for the three non-CC pathways overlapped.

Variability (measured by CV) differed systematically across pathways. In particular, the human-to-chatbot influence was highly variable due to its brief half-life producing strong but transient spikes in influence that decayed rapidly. By contrast, the chatbot self-influence had particularly low variability.

Examining the complementary cumulative distribution functions revealed differences in the proportion of relevant target-speaker messages for which influence was elevated for both the self-influence paths and the cross-agent influence paths. Most notably, influence from the chatbot to itself remained at high levels for a substantially greater proportion of chatbot messages than influences along other paths did for their respective target-speaker messages (see Table~\ref{tab:beta_z_descriptives}). On the chatbot's self-influence path, the odds of delusion were elevated by at least $1.65\times$ (i.e., $\text{log-odds} > 0.5$) during almost half of chatbot messages, and by at least $7.39\times$ (i.e., $\text{log-odds} > 2.0$) during roughly one-quarter of chatbot messages. By contrast, on the human's self-influence path, the odds of delusion were elevated by at least $1.65\times$ during 28.3\% of human messages and by at least $7.39\times$ during only 2.8\% of human messages.

On the cross-agent paths, chatbot $\to$ human influence tended to exceed human $\to$ chatbot influence except in the tail of the distribution. For example, influence exceeded 0.5 log-odds on the chatbot $\to$ human path for 30.3\% of human messages compared with 12.5\% of chatbot messages on the human $\to$ chatbot path. At the tail of the distribution, influence exceeded 2.0 log-odds for 3.2\% of chatbot messages along the human $\to$ chatbot path compared with 0.8\% of human messages on the chatbot $\to$ human path.

Taken together, these descriptive statistics of $\beta_m z_m$ indicate that chatbot self-influence was not only stronger than other paths on average, but also more stable over conversations, remaining at elevated levels for substantial portions of the dialogue. By contrast, human-to-chatbot influence was the most intermittent and volatile, characterized by high variability and lower levels of realized influence on average. This convergence of mean, variability, and tail-distribution evidence supports the overarching conclusion that humans tended to trigger sharper immediate shifts, whereas chatbots tended to maintain influence over longer conversational stretches.

\begin{table*}[t]
\centering
\begingroup
\setlength{\tabcolsep}{5pt}
\begin{tabular}{lllllllll}
\toprule
Path & Mean & 95\% CI & CV & \% > 0 & \% > .25 & \% > .5 & \% > 1.0 & \% > 2.0 \\
\midrule
CH: Reinforcement & 0.38 & [0.21, 0.48] & 1.44 & 81.2\% & 37.9\% & 30.3\% & 14.2\% & 0.8\% \\
HH: Entrenchment & 0.39 & [0.24, 0.63] & 1.59 & 79.2\% & 36.0\% & 28.3\% & 15.9\% & 2.8\% \\
HC: Mirroring & 0.24 & [0.11, 0.48] & 2.37 & 74.4\% & 18.5\% & 12.5\% & 11.8\% & 3.2\% \\
CC: Self-consistency & 1.09 & [0.77, 1.40] & 1.24 & 79.3\% & 51.7\% & 47.1\% & 39.0\% & 25.8\% \\
\bottomrule
\end{tabular}

\endgroup
\caption{Descriptive statistics of model-implied latent influence $\beta_m z_m$.}
\label{tab:beta_z_descriptives}
\vspace{0.5em}

\begin{minipage}{\textwidth}
\footnotesize
\textit{Note.} Each pathway was evaluated at messages produced by its target speaker: human messages for the $CH$ and $HH$ pathways and chatbot messages for the $HC$ and $CC$ pathways. At each such message, $\beta_m z_m$ is the pathway's model-implied contribution to the log-odds of delusion endorsement. The 95\% confidence intervals for the means were obtained by resampling participants with replacement (2000 bootstrap samples). CV is the coefficient of variation. Columns labeled $\% > x$ give the percentage of target-speaker messages for which $\beta_m z_m$ exceeded $x$. For interpretation, values of $.25$, $.50$, $1.0$, and $2.0$ correspond to $1.28\times$, $1.65\times$, $2.72\times$, and $7.39\times$ increases in the odds of delusion, respectively.
\end{minipage}
\end{table*}

\paragraph{\textit{Persistence of Latent Influence $\beta_m z_m$}}

\begin{figure}[ht]
  \centering
  \begin{minipage}{\linewidth}
    \centering
    \includegraphics[width=0.8\linewidth]{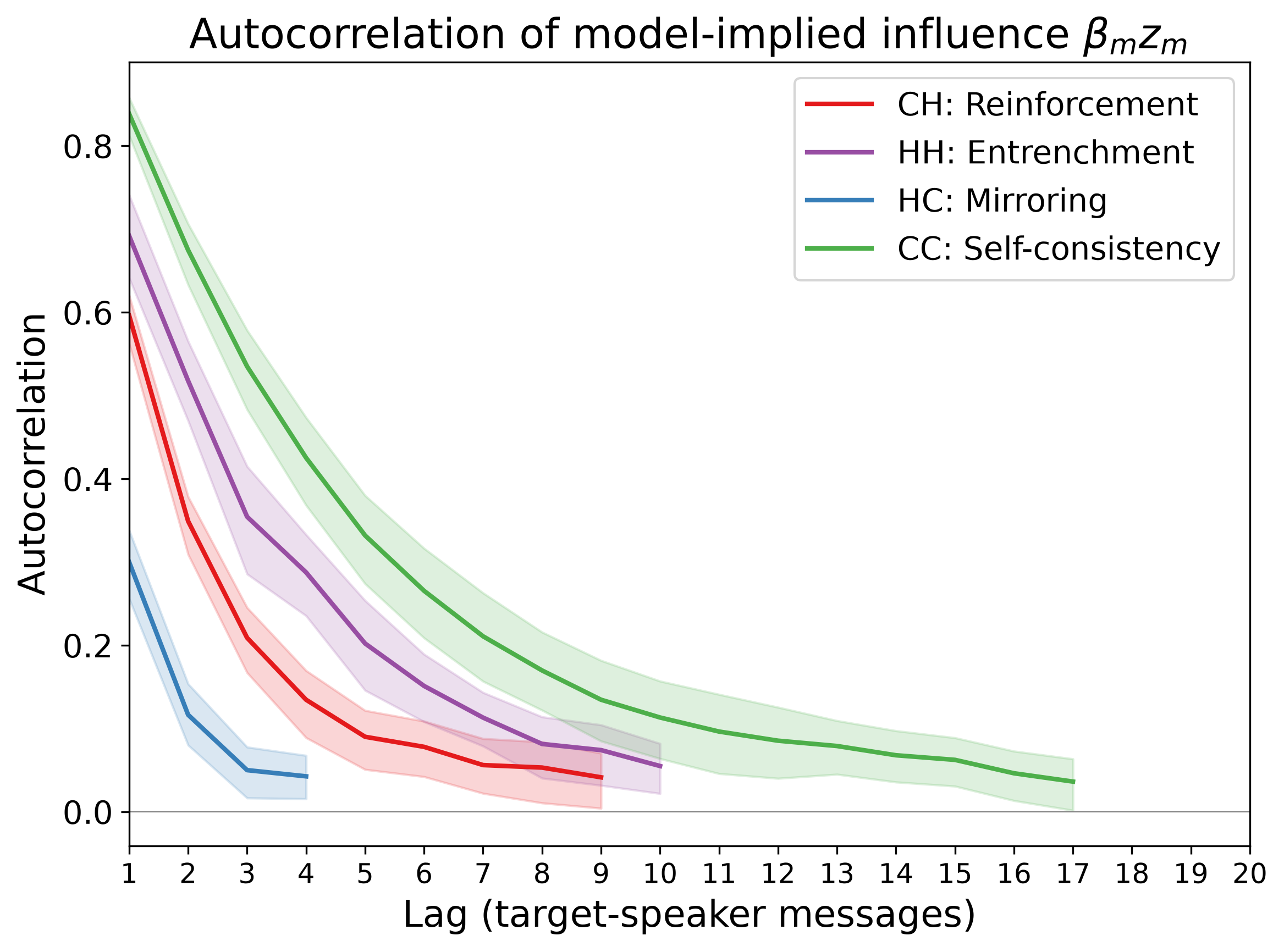}
  \end{minipage}

  \vspace{0.5em}

  \caption{Autocorrelation of model-implied pathway influence $\beta_m z_m$ across lags of 1--20 target-speaker messages. Conversation-level autocorrelations were averaged by pathway for conversations in which either speaker produced at least three delusion endorsements. Shaded regions are 95\% confidence intervals from a participant-cluster bootstrap (2000 samples). Each line ends immediately before the first lag whose confidence interval includes zero. The intervals exclude zero through lags 17 (CC), 10 (HH), 9 (CH), and 4 (HC).}
  \label{fig:acf}
\end{figure}

To quantify how long model-implied influence states persisted over the course of conversations, we examined the autocorrelation functions (ACFs) of the realized latent influence terms ($\beta_m z_m$) across lags of 1--20 target-speaker messages (see Figure \ref{fig:acf}). Higher positive autocorrelation indicates that influence present at one target-speaker message tends to remain elevated across subsequent target-speaker messages, whereas rapid decay toward zero indicates more transient dynamics.

The chatbot's self-influence pathway showed the most persistent temporal dependence: its 95\% bootstrap confidence interval excluded zero through lag 17. The intervals for human self-influence and chatbot-to-human influence excluded zero through lags 10 and 9, respectively, whereas the interval for human-to-chatbot influence excluded zero only through lag 4.

These results indicate that chatbot-originating influence appears to operate on a longer timescale (relative to human-originating influence) when comparing both the self-influence paths and cross-agent paths to one another.

\section{Discussion}

As AI systems become increasingly embedded in everyday life, they introduce both new capabilities and new risks that require systematic study. One emerging concern is the role of AI chatbots in the formation and maintenance of delusional beliefs. Disturbing reports of chatbot users who became progressively unmoored from reality, seemingly by the relentless encouragement of chatbots, have proliferated in the media and in some cases escalated into court proceedings and legislative scrutiny. It remains unclear, however, whether and how the behavior of the chatbot played a role in the development of such delusional episodes.

In our work, we (1) examined whether delusional human--chatbot conversations show evidence of bidirectional false belief amplification and (2) decomposed the process into four influence pathways and characterized the temporal dynamics of each pathway. \textbf{Across 19 participants and over 390,000 messages, we found evidence consistent with the hypothesis that delusional beliefs were sustained and amplified through positive feedback loops between humans and chatbots.} Though people and chatbots both contributed to the shared delusions, the two parties exerted asymmetric effects on one another. People provided strong but short-lived influences on the chatbot, while the chatbot exerted longer-lasting, persistent influence on the people. However, the most dominant influence pathway carrying delusions throughout the conversational thread was the chatbot's influence on itself. The chatbot's self-influence pathway exhibited the greatest aggregate magnitude of influence of the four possible pathways and influence along this path tended to remain elevated for long periods once it had become elevated. In sum, this pattern suggests that \textbf{people may be potent in the moment in instantiating a delusion, but the chatbot acts as a flywheel that perpetuates the delusion over time}.

\subsection{Theoretical Implications}

Our findings have several implications for understanding how human-chatbot delusional conversations develop. First, they provide initial empirical support for a bidirectional model of belief amplification rather than a unidirectional one.

Second, viewed alongside prior work on delusional ideation, our results suggest that chatbots may increase the clinical severity of delusional beliefs by amplifying preoccupation and conviction. Although delusional beliefs are relatively common in the general population, clinically significant delusions are distinguished less by their presence than by greater preoccupation, conviction, and distress \citep{petersMeasuringDelusionalIdeation2004}. By sustaining and reinforcing delusional content over time, chatbots may increase preoccupation (persistence of the belief) and conviction (strength of the belief), potentially shifting otherwise benign delusional ideation toward clinically significant delusions. This interpretation is consistent with the stress-vulnerability model, which has recently been proposed as an explanation for ``AI psychosis'' in vulnerable individuals \citep{hudon_delusional_2025}.

Third, chatbot self-consistency (chatbot $\to$ chatbot influence) was the strongest pathway sustaining delusional content. Public discussion has largely attributed AI-fueled delusions to sycophancy, which we decompose into belief reinforcement (chatbot $\to$ human) and belief mirroring (human $\to$ chatbot). Although both pathways contributed to belief transmission, the chatbot's tendency to remain consistent with its own prior responses was substantially more influential. This finding is notable given recent work devoted to improving LLM self-consistency \citep{abdulhaiConsistentlySimulatingHuman2025, presPositionItsTime}. Our results suggest that self-consistency is a double-edged sword: when a conversation becomes delusional, maintaining consistency can propagate and amplify the chatbot's own erroneous outputs, which subsequently reinforce the user's beliefs through later chatbot responses.

\subsection{Implications for Chatbot Evaluation and Safeguards}

Our work also has implications for chatbot evaluation and moderation. First, the dominance of the chatbot self-consistency pathway (CC) suggests that evaluations should assess not only whether a chatbot avoids generating harmful messages, but also whether it can recover after producing one. Existing evaluations typically simulate a delusion-prone user and assess the chatbot's response. While this is important, recovery from prior unsafe outputs is a distinct capability. Given the billions of chatbot messages sent each week, some harmful responses are inevitable, making robust recovery mechanisms essential \citep{openai2026scaling}.

Second, our findings suggest that evaluations and safeguards should detect and interrupt gradual belief escalation across turns rather than focusing only on single-message failures. The accumulation of latent influence we observed indicates that harm often emerges through sustained reinforcement rather than isolated responses. Consistent with this, some conversations in our dataset appeared to begin as role-play or fiction before gradually evolving into authentic delusional belief for both the human and chatbot \citep{moore2026delusional, yang_ai-induced_2026}.

Current evaluation methods vary in how well they capture this process. Single-turn benchmarks miss the temporal dynamics of belief escalation \citep{shenEvaluationLargeLanguage}. Multi-turn benchmarks with simulated users better capture these dynamics but rely on implicit theories of user behavior \citep{kirgisLLMSpiralsDelusion2026, paech_spiral-bench_2025}. More recent approaches encode explicit theories of belief change through prompting \citep{yeung_psychogenic_2025}. Our results suggest a next step: grounding evaluations in real human-chatbot delusional conversations, either by replaying authentic conversation trajectories or by training user simulators on empirical conversation data.

Finally, our work demonstrates a general framework for modeling the accumulation and decay of latent influence states from ordered dialogue annotations. Because model fitting requires binary annotations rather than transcript text, similar approaches could reduce the exposure of raw text during aggregate evaluation and moderation for risks including delusions, mental health crises, and relational over-attachment, using annotation pipelines such as CLIO \citep{tamkin_clio_2024}. 

\subsection{Concluding Remarks}

Our findings provide the first quantitative empirical support for an account of bidirectional belief amplification in chatbot conversations where false beliefs emerge from the dynamics of mutual influence between humans and AI chatbots over the course of extended conversations. As people increasingly participate in long, often deeply personal conversations with AI chatbots, understanding these dynamics through the lens of latent cognitive processes will be essential for mitigating harms and supporting human flourishing.

\section*{Limitations}

Our work provides a critical initial step in understanding how long-context delusional conversations unfold between people and chatbots. However, several important limitations and opportunities for future work should be noted.

First, findings in our report, like all work with human participants, must be considered conditional on our dataset. Our dataset consists of 19 users of AI chatbots who experienced relatively severe delusional episodes and submitted their chat transcripts to a study on chatbot harms. Therefore, the findings in our report are not considered universally true across all people or even all chatbot users. That said, given the growing (and already immense) prevalence of chatbot use, even subpopulations that constitute a very small proportion of the total population can include thousands or hundreds of thousands of people \citep{openai_strengthening_2025}.

Second, while temporal dependence provides more certainty than anecdotal reports or even cross-sectional statistical correlations, we are limited in our inferences by the lack of an experimental manipulation. Since it would be unethical to evoke extended delusional episodes like those observed in our study, we used temporal dependence as the best approximation to more rigorous causal inference methods.

Third, our work modeled the influence of bot \textit{endorsements} of delusions. However, work should be done on the impact of bot \textit{refusals and corrections} in delusional conversations to additionally inform strategies for mitigating delusional thinking in chatbot users.

Fourth, the binary endorsement labels contain measurement error. Although we report validation against human judgments, the model treats annotations as observed without error and does not propagate uncertainty from the LLM classifier. The labels capture message content, not latent belief, conviction, distress, or clinical state. The chatbot labels likewise describe output rather than a chatbot internal state.

\section*{Ethical Considerations}

This work analyzes highly sensitive conversations donated by people who reported harm from AI chatbots, including experiences described as delusional. The study therefore poses privacy, dignity, and stigmatization risks beyond those of ordinary dialogue research. All participants provided informed consent, and all study procedures received institutional review board approval. Transcripts were processed with Microsoft Presidio to remove names, locations, and other identifying details. We do not release raw transcript text, timestamps, source-file information, or original identifiers; the released event data contain only binary message annotations needed to reproduce the aggregate analyses. 

``Delusion endorsement'' is a research annotation of message content, not a clinical diagnosis. Diagnosis requires additional context and qualified assessment, and culturally grounded, religious, figurative, fictional, or joking statements may be especially vulnerable to misclassification. Automated labeling could stigmatize people, trigger inappropriate interventions or account restrictions, intensify surveillance, or impose unequal burdens on groups underrepresented in the annotation process. Given the classifier's imperfect accuracy and the selected sample, the present pipeline is not appropriate for diagnosis, clinical triage, population screening, or automated enforcement.

The intended benefit of this work is to support research and safety evaluation of escalating, harmful chatbot interaction patterns. There are also misuse risks: similar methods could be used to profile vulnerable users, exploit moments of vulnerability, or optimize engagement with harmful conversations. The findings could also be overgeneralized or misread as causal evidence, distorting clinical or policy decisions. Any downstream safety system motivated by this work should protect privacy, include meaningful human review and avenues for contesting decisions, involve affected communities and mental-health expertise, and evaluate disparate errors and unintended harms before deployment.

\section*{Acknowledgments}

This work was made possible by generous funding from the Yidan Prize Foundation to C.D. and the National Science Foundation under Award No. 2443038 to D.C.O. This work was supported by Gemini API credit grants through the Stanford Institute for Human-Centered Artificial Intelligence (HAI). Any opinions, findings and conclusions or recommendations expressed in this material are those of the authors and do not necessarily reflect the views of our funders.

We also thank Andreas Haupt and Kate Petrova for helpful conversations and feedback on the manuscript.

We used AI assistants (GPT/Codex) for manuscript review and editing and for agentic AI coding. The authors reviewed all AI-generated content for correctness and retain responsibility for the final manuscript, analyses, and claims.

\bibliography{custom, anthis_zotero_sync, manual}

\clearpage
\appendix

\section{Detailed Model Specification}
\label{sec:appendix}

\subsection{Model Specification}
\label{sec:model-specification}

Let $j$ index participants, $c$ index conversations, and $S,T \in \{C,H\}$ denote the source and target speakers, respectively, where $C$ is the chatbot and $H$ is the human participant. Within each conversation, $t$ indexes messages on the target speaker's own message grid rather than on a joint grid containing both speakers. Thus, $D_{H,t,c,j}$ indicates whether the $t$th human message endorses a delusion, and $D_{C,t,c,j}$ analogously denotes the $t$th chatbot message. Subscripts list the source speaker first and target speaker second: $CH$ is chatbot-to-human influence, $HC$ is human-to-chatbot influence, and $CC$ and $HH$ are chatbot and human self-influence, respectively. Let $m=ST \in \{CH,HC,CC,HH\}$ index these four pathways.

For each target-speaker message, we model the probability of delusion endorsement conditional on the preceding conversation history, $\mathcal{F}$, using a logistic link. Writing $\eta_{S,t,c,j}=\operatorname{logit}\Pr(D_{S,t,c,j}=1\mid\mathcal{F}_{S,t,c,j})$, the two observation models are

\begin{align}
\eta_{H,t,c,j}
&= \alpha_{H,j} \notag \\
&\quad + \beta_{CH,j}\, z_{CH,t,c,j} \notag \\
&\quad + \beta_{HH,j}\, z_{HH,t,c,j}
\label{eq:human_logit} \\
\eta_{C,t,c,j}
&= \alpha_{C,j} \notag \\
&\quad + \beta_{HC,j}\, z_{HC,t,c,j} \notag \\
&\quad + \beta_{CC,j}\, z_{CC,t,c,j}
\label{eq:chatbot_logit}
\end{align}

where $z_m$ represents the latent influence states described below. The $\alpha$ and $\beta$ parameters are normally distributed participant-specific random intercepts and slopes with means and variances

\begin{align}
\alpha_{H,j} &\sim \mathcal{N}(\mu_{\alpha_H}, \sigma_{\alpha_H}^2), 
& \alpha_{C,j} &\sim \mathcal{N}(\mu_{\alpha_C}, \sigma_{\alpha_C}^2) 
\end{align}

\begin{align}
\beta_{m,j} &\sim \mathcal{N}(\mu_{\beta_m}, \sigma_{\beta_m}^2)
\end{align}

\paragraph{\textit{Latent Influence States ($z_{m,t,c,j}$) and Impulses ($u_{m,t,c,j}$)}}

Each latent state evolves on the message grid of its target speaker $T$. For a cross-agent pathway $m=ST$, let $\mathcal{G}_{m,t,c,j}$ be the set of source-speaker turns occurring after target speaker $T$'s preceding message and before $T$'s current message; for $t=1$, the interval begins at the conversation boundary. The pathway-specific impulse supplied before target turn $t$ is
\begin{equation}
u_{m,t,c,j} =
\begin{cases}
0,
& S=T,\ t=1, \\
D_{T,t-1,c,j},
& S=T,\ t>1, \\
\displaystyle\sum_{r\in\mathcal{G}_{m,t,c,j}}D_{S,r,c,j},
& S\ne T.
\end{cases}
\label{eq:u-def}
\end{equation}
For a self-influence pathway, the impulse is therefore the delusion indicator for the target speaker's immediately preceding message. For a cross-agent pathway, it is the sum of all delusion-endorsing source-speaker messages that occurred after the preceding target-speaker message and before the current one. Consequently, multiple source messages between two target turns contribute additively to the same impulse.

The latent state then follows
\begin{equation}
z_{m,t,c,j}
= \mathrm{e}^{-\lambda_m}
\big(z_{m,t-1,c,j}+u_{m,t,c,j}\big).
\label{eq:latent-update}
\end{equation}
The initial state is $z_{m,0,c,j}=0$. Thus, decay advances once per target-speaker turn, not once per message on the joint human--chatbot timeline. All latent states are reset to zero at conversation boundaries.

\paragraph{\textit{Decay Parameters ($\lambda_m$)}}

Each pathway has its own decay rate parameter, which governs how rapidly influence from prior message history dissipates over target-speaker turns. For ease of interpretation, we convert the estimated decay rates into half-lives (Eq.~\ref{eq:lambda-dist}), which represent how many turns of the target speaker---the speaker whose behavior is being predicted---it takes for 50\% of the influence to dissipate. This conversion is for interpretability and does not affect model estimation. Specifically,

\begin{equation}
\label{eq:lambda-dist}
\quad h_m = \ln(2)/\lambda_m.
\end{equation}

\subsection{Adaptive Gauss--Hermite Model Fitting}
\label{sec:aghq}

Let $\mathbf{b}_{T,j}=(\beta_{ST,j},\beta_{TT,j},\alpha_{T,j})^\top$ denote the three random effects for participant $j$ in target-speaker block $T$, let $\boldsymbol{\theta}_T$ collect that block's population-level parameters, and let
\begin{equation}
\begin{aligned}
g_{T,j}(\mathbf{b};\boldsymbol{\theta}_T)
={}&-\log p(\mathbf{D}_{T,j}\mid\mathbf{b},
\boldsymbol{\theta}_T) \\
&-\log p(\mathbf{b}\mid\boldsymbol{\theta}_T)
\end{aligned}
\label{eq:joint-nlp}
\end{equation}
be the participant-level joint negative log density. The first term is the sum of Bernoulli-logistic negative log likelihoods over all target-speaker messages from participant $j$; the second is the negative log density of the three independent Gaussian random effects. The participant's contribution to the marginal likelihood is
\begin{equation}
L_{T,j}(\boldsymbol{\theta}_T)
=\int_{\mathbb{R}^3}
\exp\{-g_{T,j}(\mathbf{b};\boldsymbol{\theta}_T)\}\,d\mathbf{b}.
\label{eq:marginal-integral}
\end{equation}

At every evaluation of the outer objective, we found the conditional mode
$\widehat{\mathbf{b}}_{T,j}=\arg\min_{\mathbf{b}}g_{T,j}(\mathbf{b};\boldsymbol{\theta}_T)$
by Newton iterations with backtracking. We stopped the inner solve when the largest absolute gradient component was below $10^{-9}$, allowing at most 60 iterations. Let $\mathbf{H}_{T,j}$ be the Hessian of $g_{T,j}$ at the mode and let $\mathbf{A}_{T,j}\mathbf{A}_{T,j}^{\top}=\mathbf{H}_{T,j}^{-1}$ be its Cholesky factorization. We added $10^{-6}\mathbf{I}$ to the conditional Hessian for numerical stability. For a $d$-dimensional tensor-product Gauss--Hermite rule with nodes $\mathbf{x}_q$ and product weights $w_q$, the adaptive nodes are
\begin{equation}
\mathbf{b}_q
=\widehat{\mathbf{b}}_{T,j}
+\sqrt{2}\,\mathbf{A}_{T,j}\mathbf{x}_q,
\end{equation}
and the integral is approximated as
\begin{equation}
\begin{aligned}
\widehat{L}_{T,j}
={}&2^{d/2}|\mathbf{H}_{T,j}|^{-1/2} \\
&\times\sum_q w_q
\exp\{-g_{T,j}(\mathbf{b}_q;\boldsymbol{\theta}_T) \\
&\hspace{72pt}+\lVert\mathbf{x}_q\rVert^2\}.
\end{aligned}
\label{eq:aghq}
\end{equation}
For the bidirectional model, $d=3$ and the five-node-per-dimension rule contains 125 points for each participant in each target-speaker block. The adaptive transformation was recomputed separately for every participant and every candidate value of $\boldsymbol{\theta}_T$.

We minimized $-\sum_j\log\widehat{L}_{T,j}$ separately for the human-target and chatbot-target blocks using bounded L-BFGS-B. Three-point finite differences supplied the outer gradient, with relative steps of $10^{-3}$ for the human-target block and $10^{-4}$ for the chatbot-target block; the optimizer used a gradient tolerance of $10^{-6}$ and a relative function tolerance of $10^{-13}$. Both blocks were cold-started with decay rates of $0.2$, random-slope means of $0$, random-slope standard deviations of $\exp(-1)$, an intercept mean of $0$, and an intercept standard deviation of $1$. Positive parameters were represented on the log scale. We evaluated quadrature accuracy by recomputing each final objective with seven nodes per dimension while holding the fitted parameters fixed. Finally, we computed the Hessian of the five-node marginal negative log likelihood by centered finite differences with relative step $10^{-3}$; its inverse supplied the covariance matrix used to construct 95\% Wald confidence intervals.

The unidirectional lesion was fit identically except that fixing the chatbot-to-human pathway to zero reduced the human-target random-effects vector to $(\beta_{HH,j},\alpha_{H,j})^\top$. Its AGHQ rule therefore had $d=2$ and 25 points per participant. For the likelihood-ratio comparison, we retained the Gaussian normalization constants in both models because their random-effects dimensions differ.

\end{document}